\documentclass{article}
\usepackage{ijcai23}

\usepackage{fancyhdr}
\usepackage{times}
\usepackage{soul}
\usepackage{url}
\usepackage[utf8]{inputenc}
\usepackage[small]{caption}
\usepackage{graphicx}
\usepackage{amsmath}
\usepackage{amsthm}
\usepackage{booktabs}       
\usepackage{algorithm}
\usepackage{algorithmic}
\usepackage[switch]{lineno}

\usepackage[utf8]{inputenc} 
\usepackage[T1]{fontenc}    
\usepackage{amsfonts}       
\usepackage{nicefrac}       
\usepackage{microtype}      
\usepackage{xcolor}         
\usepackage{enumitem}
\usepackage{caption}
\setlist{topsep=2pt, itemsep=1pt, leftmargin=5.5mm}
\usepackage{natbib}
\defcitealias{fred_2022_wti}{U.S. EIA, 2022}
\defcitealias{fred_2022_brent}{U.S. EIA, 2022}

\makeatletter
\renewcommand\hyper@natlinkbreak[2]{#1}
\makeatother

\usepackage{siunitx} 
\usepackage[colorlinks=true, citecolor=violet, linkcolor=red, urlcolor=blue]{hyperref}

\title{Real World Time Series Benchmark Datasets with Distribution Shifts:\\Global Crude Oil Price and Volatility}

\author{
    Pranay Pasula\footnotemark[1]\footnotemark[2]\footnotemark[3]
    \affiliations
    UC Berkeley EECS
    \emails
    \texttt{pasula@berkeley.edu}
}
\begin{document}

\maketitle

\footnotetext[1]{Accepted to IJCAI 2023, the 32nd International Joint Conference on Artificial Intelligence (AI4TS).}
\footnotetext[2]{Selected as a Contributed Talk.}
\footnotetext[3]{Awarded Best Paper Runner-Up / Honorable Mention.}

\thispagestyle{plain}

\begin{abstract}
    The scarcity of task-labeled time-series benchmarks in the financial domain hinders progress in continual learning. Addressing this deficit would foster innovation in this area. Therefore, we present \textbf{COB}, \textbf{C}rude \textbf{O}il \textbf{B}enchmark datasets. COB includes 30 years of asset prices that exhibit significant distribution shifts and optimally generates corresponding task (i.e., regime) labels based on these distribution shifts for the three most important crude oils in the world. Our contributions include creating real-world benchmark datasets by transforming asset price data into volatility proxies, fitting models using expectation-maximization (EM), generating contextual task labels that align with real-world events, and providing these labels as well as the general algorithm to the public. We show that the inclusion of these task labels universally improves performance on four continual learning algorithms, some state-of-the-art, over multiple forecasting horizons. We hope these benchmarks accelerate research in handling distribution shifts in real-world data, especially due to the global importance of the assets considered. We've made the (1) raw price data, (2) task labels generated by our approach, (3) and code for our algorithm available at \href{https://oilpricebenchmarks.github.io}{oilpricebenchmarks.github.io}.
\end{abstract}

\section{Introduction}
\label{sec:introduction}

Benchmarks serve as testbeds for artificial intelligence (AI) algorithms and have promoted the rapid acceleration of algorithm performance that has been seen over the past ten years. Notable examples from the beginning of this ten year period include ImageNet \citep{imagenet}, COCO \citep{coco}, CIFAR-10 \citep{krizhevsky2009learning}, and the Arcade Learning Environment \citep{bellemare2013arcade}, the first of which is often credited with initiating the ongoing deep learning revolution.

These benchmarks all contain images in the form of raw pixel data, which is an important medium, but more recently, benchmarks have expanded to include different modalities of data that are aimed at evaluating progress in a variety of AI subfields and problem domains. 

In this work, we focus on \emph{temporal distribution shifts} within the financial domain and introduce time-series benchmark datasets of the three primary global crude oil price markers, \emph{West Texas Intermediate (WTI)}, \emph{Brent Blend}, and \emph{Dubai Crude}. 

For each dataset we provide over 30 years of data---daily spot prices for WTI and Brent and monthly average prices for Dubai Crude---and see striking distribution shifts throughout this time period for each asset.


Problem settings with time-series or otherwise sequential data often pose serious issues to predictive AI models in several ways that the standard supervised learning setting using (independent and identically distributed) IID non-sequential data does not. Time-series often embodies obstacles, such as 

\begin{itemize}
    \item \textbf{Correlatedness}. While training predictive models, series data points that occur near the same time tend to take similar values and have similar targets. This biases the models to this local region of the input and target variable spaces.
    \item \textbf{Non-stationarity}. Detailed in Section \ref{sec:non-stationarity}.
    \item \textbf{Missing data}. Subsets of data are often missing from series data. For example, inadequate record keeping may be why the WTI and Brent datasets we introduce are missing 3 and 2 percent of daily spot price data, respectively, but we resample the datasets in a way that results in new series without missing data. 
    \item \textbf{Spurious data}. Outliers, erroneous, or infeasible data skew model learning in poor directions.
\end{itemize}

\subsection{Non-Stationarity}
\label{sec:non-stationarity}
Non-stationarity, or distribution shifts, threaten predictive models that have been trained on data and then deployed to make predictions online, or in real-time on different data. As the data distribution shifts further from the data distribution the model was trained on, the worse the model performs. Furthermore, this non-stationarity can manifest in fundamentally different ways. 

Let $x$ be the independent variables, $y$ be the target variables, and $P$ be an appropriate probability measure, then we have the following types of non-stationarity:

\begin{itemize}
    \item \textit{Covariate shift}: A shift in the distribution of independent variables $P(X)$.
    \item \textit{Prior probability shift}: A shift in the distribution of target variables $P(Y)$.
    \item \textit{Concept drift}: A shift in the distribution of the relationship between the independent and target variables $P(Y \mid X)$.
\end{itemize}

\subsection{Out-of-Distribution Detection}
\label{sec:oodd}
It's crucial for online algorithms to quickly identify distribution shifts because predictive models perform worse as the data they are predicting on deviates further from the data that they were trained on. \emph{But how does an algorithm decide that a distribution shift has occurred?} There's a trade-off between speed and confidence in deciding whether a recent sequence of data includes a distribution shift.  


\subsection{Continual (Lifelong) Learning}
\label{sec:continuallearning}
A rapidly growing field of artificial intelligence that often handles data online is \emph{continual learning}, which is sometimes used interchangeably with \emph{lifelong learning} or \emph{incremental learning}. Continual learning algorithms aim to develop models that accrue useful knowledge over time to accomplish some set of tasks, or equivalently, perform well on data distributions corresponding to some set of underlying tasks while avoiding \textit{negative interference} or \textit{catastrophic forgetting}.

We refer to the typical continual learning problem setting as one in which the tasks are known, regime shifts happen instantaneously during a discrete timestep, and both the regime-generated data and the regime labels are given to the model at any time $t$.

\subsection{Our Contribution}
\label{subsec:contribution}
Our main contribution is the creation of three new real-world benchmark datasets on prices of WTI, Brent, and Dubai crude oil, which are of critical importance worldwide \citep{backus2000oil, kilian2009impact}. 

We have 
\begin{enumerate}
\item \textbf{Transformed} the data into a proxy for volatility,
\item \textbf{Fitted} the data into tasks using expectation-maximization (EM), a well-established algorithm, to optimize for measures of information criteria,
\item \textbf{Provided} contextual labels of major real-world events that align with price and regime shifts,
\item \textbf{Demonstrated} that our algorithmically task-labeled dataset generated tasks that align with recessions and other spurious real-world events.
\end{enumerate}

We hope that this work will accelerate the development of continual learning, out-of-distribution detection, and other algorithms in handling issues posed by real-world data that contains distribution shifts. Furthermore, we believe that the societal importance of the assets represented by these datasets makes these benchmarks especially appealing.




\section{Datasets}

A description of WTI, Brent, and Dubai Crude oils can be found in Section \ref{sec:datasets}.

The raw data for WTI and Brent contains prices at a daily frequency while that for Dubai Crude contains average prices at a monthly frequency. Therefore the resampling process described in Section \ref{sssec:appropriate_resampling} applies only to WTI and Brent.

\section{Dataset Creation Process}


This section elucidates the steps that we have taken to construct the real-world benchmark datasets that we are providing.

\subsection{Converting Daily Spot Price to Volatility}

\subsubsection*{WTI and Brent}
We use the daily spot prices of WTI \citepalias{fred_2022_wti} and Brent \citepalias{fred_2022_brent} to derive weekly percent changes, a proxy of volatility, which is important in evaluating \emph{value at risk} (VaR), a statistic that quantifies the extent of potential financial losses within a position over a certain period of time.

\subsubsection*{Dubai Crude}
We use the monthly average prices of Dubai Crude \citep{fred_2022_dubai} to derive monthly percent changes, another proxy of volatility, but one that is attenuated by the averaging over the monthly prices, as seen by comparing Figure \ref{fig:dubai_to_labels} against Figure \ref{fig:wti_to_labels} and Figure \ref{fig:brent_to_labels}.

\subsubsection*{Heterogenous Dataset Difficulty}
The difference in pricing frequency between the first two, WTI and Brent, and the last, Dubai, allow us to create benchmark datasets with different difficulties. Since Dubai Crude prices are monthly averages, much of the pricing spread is smoothed out relative to if daily spot prices were used. Unfortunately, the most reliable and granular Dubai Crude prices that had history comparable in length to the WTI and Brent prices we use are the monthly average prices we source from \citep{fred_2022_dubai}.

However, we see this as an opportunity to present benchmark datasets with varying degrees of difficulty on various tasks, most obviously to make predictions on, as target variables with lower spread are easier for predictive models to handle.

\subsection{Selecting Model Class Used to Create Tasks}

\subsubsection*{Markov Regime Switching Models}
Markov switching models are used on sequential data that is expected to transition through a finite set of latent states, or tasks. These states can and often differ significantly, as evidenced in Figure \ref{fig:wti_to_labels}, Figure \ref{fig:brent_to_labels}, and Figure \ref{fig:dubai_to_labels}.

\citep{kim1998testing} describes a Markov switching model that is particularly well suited to handling the heteroskedasticity and mean reversion properties of the raw data. The algorithm used in \citep{kim1998testing} has additional nice properties, such as the use of expectation-maximization (EM), a form of (local) \textit{maximum a posteriori} (MAP) estimation.

Therefore, we use \citep{kim1998testing} to construct the benchmark datasets and describe these steps in Section \ref{ssec:cttfm}.

\subsection{Creating Tasks Through Fitting model class}
\label{ssec:cttfm}

\subsubsection{Appropriate Resampling}
\label{sssec:appropriate_resampling}

The raw WTI and Brent spot price data was captured daily, and the day-to-day differences oscillated between positive and negative values so frequently that the method we used to generate tasks \citep{kim1998testing} oscillated between tasks so frequently that the results were unrealistic. Resampling the data by every two days before using \citep{kim1998testing} didn't resolve the issue, as evidenced by Figure \ref{fig:2dresamp} in Appendix \ref{app:resamp}.

We found that resampling weekly was an excellent frequency for \citep{kim1998testing} to generate tasks that were reasonable and well optimized measures of likelihood and information criteria. 

\subsubsection{Evaluating Stationarity}

Since we're putting forth the WTI, Brent, and Dubai crude oil datasets as benchmark datasets, we want them to be amenable to several types of evaluations. The most common evaluation on these series data are statistical inference, specifically forecasting future asset prices. \citep{nagabandi2018deep, he2019task, caccia2020online, he2021clear, pasula2023mixture} have shown that the provision of task, or context, information to predictive models boosts their ability to accurately predict future values on a variety of problem domains with challenging properties.

Therefore, it's important to ensure that the time-series benchmark datasets we're providing are useful for evaluating tasks such as forecasting or statistical inference. A critical step towards this is verifying that this series data are not characterized by unit root processes, which can lead to issues in statistical inference on this data \citep{phillips1988testing}.

To evaluate whether unit roots are present in any of the benchmark datasets, we perform an augmented Dickey-Fuller (ADF) test on each of the datasets. The ADF tests on the datasets reject the null hypotheses that unit roots are present in the data with all p-values $< \num{1e-23}$.

\subsubsection{Choosing the Number of Tasks}
\label{sssec:task_num_choice}

The problem of choosing the number of tasks $k$ is one of choosing the model itself: once $k$ is specified, \citep{kim1998testing} uses EM to fit the data to a partition of $k$ tasks. A rigorous way to guide model selection, and hence $k$, is to use statistical measures of fit known as information criteria \citep{stoica2004model}. The Akaike, Bayesian, and Hannan-Quinn information criteria (AIC, BIC, HQIC) are three of the most commonly measures. Each penalizes model deviation the same way but they vary the degree that they penalize the number of model parameters and the loss of degrees of freedom.

We choose $k$ so that $\text{AIC} + \text{BIC} + \text{HQIC}$ is minimized under the constraint that each task is present for at least one timestep for every dataset. This results in $k=3$. We provide more information on our decision process for this in Appendix \ref{app:infocri}.

\subsubsection{Smoothing probabilities}

At any time $T$ with $t<T$, \citep{kim1998testing} uses an algorithm introduced in \citep{kim1994dynamic} to smooth conditional probabilities $P(\mathcal{T}_t=i \mid Y_t)$ for every $t$ based on all observations available up through time $T$.

\subsubsection{Assigning Tasks}

For every timestep $t$, we assign task $\mathcal{T}_t$ to the argmax of the smoothed probabilities at $t$ over all tasks $\{1, 2, \dotsc, k\}$.

\section{Datasets}
\label{sec:datasets}

\subsection{West Texas Intermediate Crude Oil Daily Spot Price}

\begin{figure*}[!htb]
  \centering
  \includegraphics[width=0.85\textwidth]{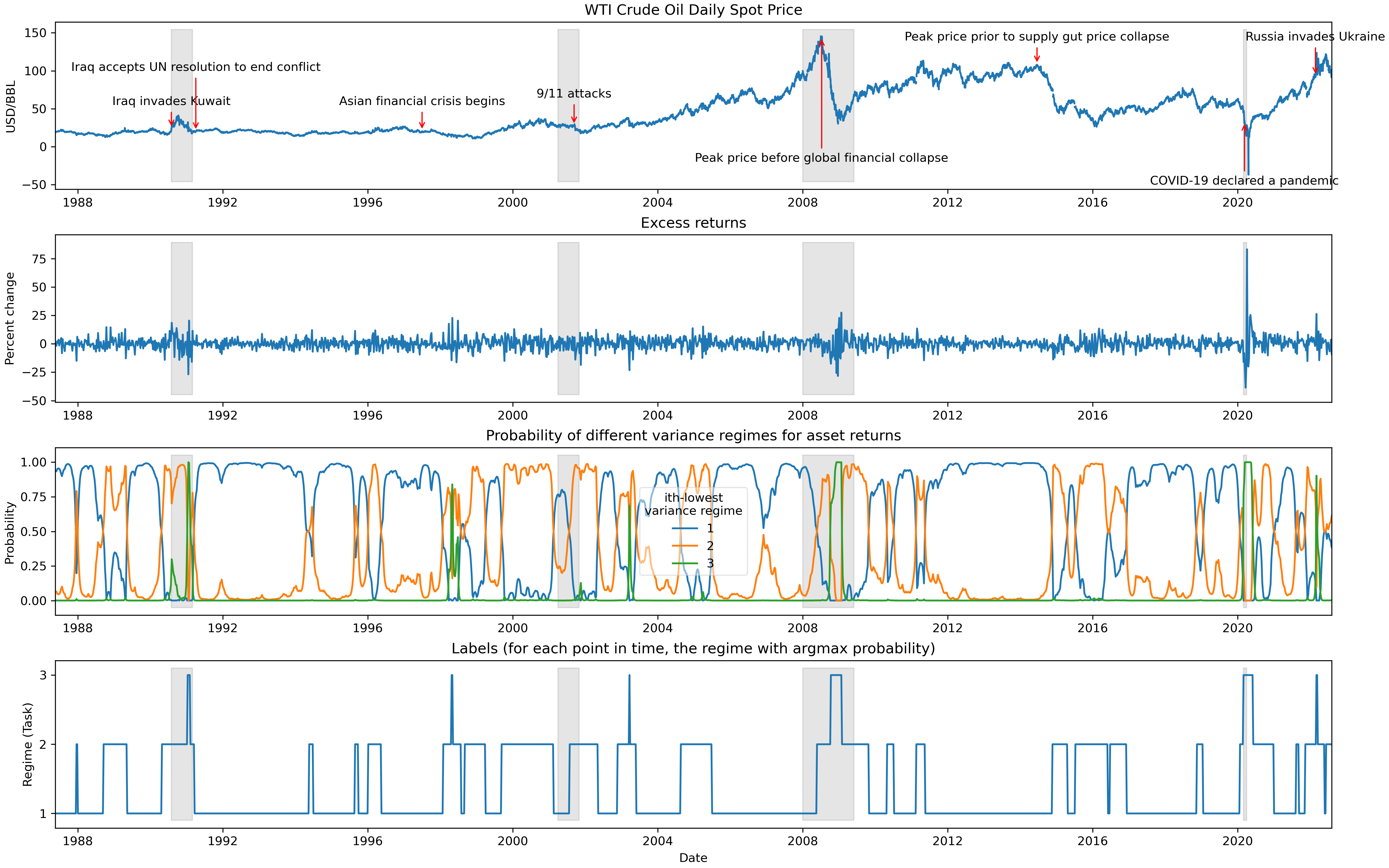}
  \captionof{figure}{Gray areas are periods that the NBER has classified as a recession. \textit{(Top)} WTI daily spot price. \textit{(Middle top)} Excess returns, a proxy of volatility and variance. \textit{(Middle bottom)} Fitted model regime probabilities over time. \textit{(Bottom)} Task labels for each point in time, computed via argmax probabilites.}
  \label{fig:wti_to_labels}
\end{figure*}

A light (low density) and sweet (low sulfur content) crude oil that is ideal for gasoline refining. This oil is extracted from land wells in the United States and is transported to a location in the center of the US, Cushing, Oklahoma. The cost required to transport WTI from its land-locked location to overseas areas of the globe is a major drawback to this otherwise high-quality crude oil and contributes to the positive Brent-WTI spread. WTI is the main benchmark for oil that is used within the United States. We depict our benchmark dataset creation algorithm for WTI in Figure \ref{fig:wti_to_labels}.

\subsection{Brent Crude Oil Daily Spot Price}

\begin{figure*}[!htb]
  \centering
  \includegraphics[width=0.85\textwidth]{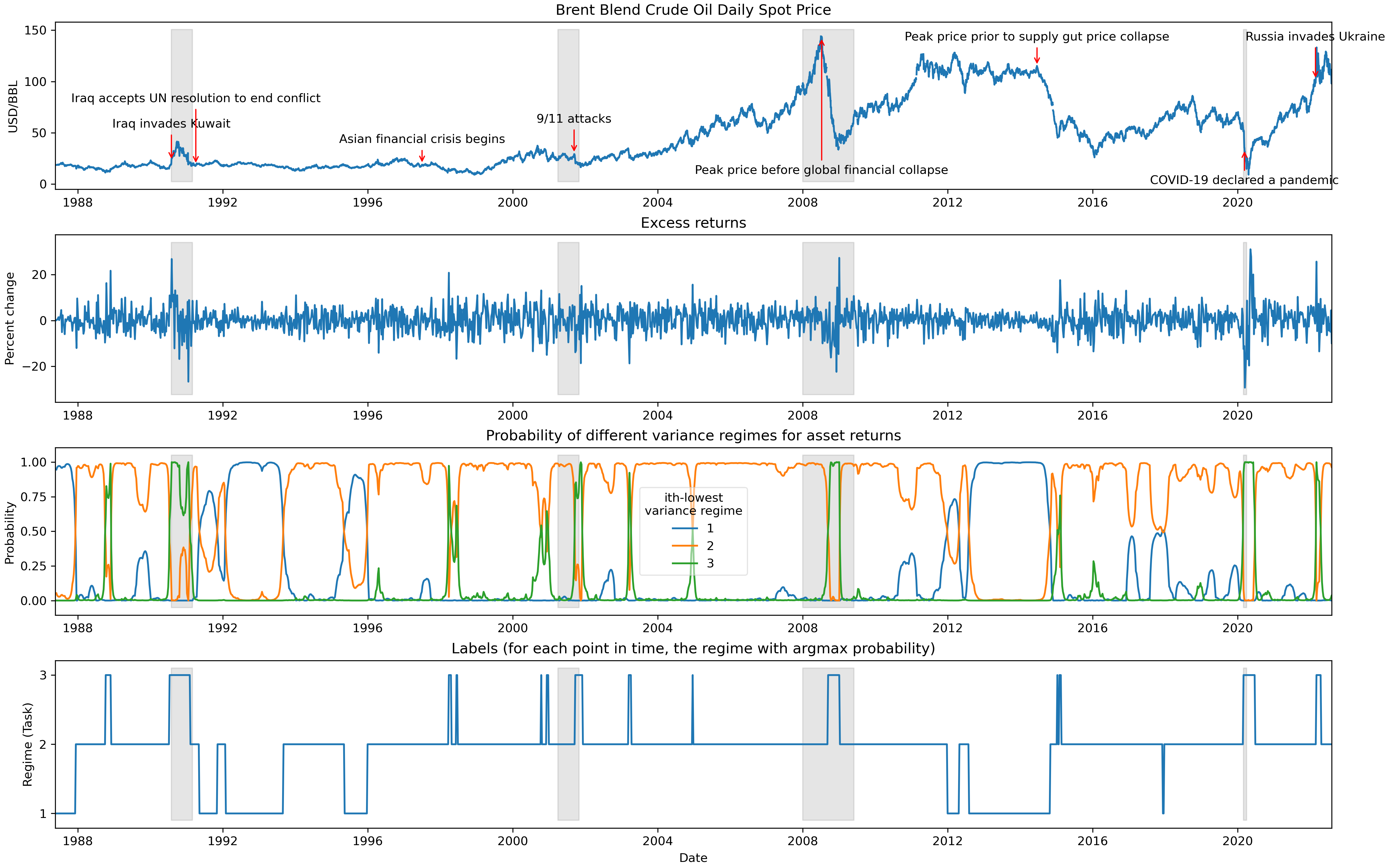}
  \captionof{figure}{\textit{(Top)} Brent crude oil daily spot price per barrel. \textit{(Middle top)} Percent change over the weekly end price, a proxy of volatility and variance. \textit{(Middle bottom)} Fitted model regime probabilities over time. \textit{(Bottom)} Regime labels for each point in time, computed via argmax probabilites.}
  \label{fig:brent_to_labels}
\end{figure*}

The most popular oil contracted, making up approximately two-thirds of all global crude contracts, "Brent" is a blend of oil from four fields in the North Sea and, like its WTI counterpart, is light and sweet. Because it can be distilled into high-value products, such as gasoline and diesel fuel, and due to it being sea-sourced and thus having relatively low transportation costs globally, it has a far greater global reach and generally higher value than WTI. We depict our benchmark dataset creation algorithm for Brent in Figure \ref{fig:brent_to_labels}

\subsection{Dubai Crude Oil Monthly Average Price}

\begin{figure*}[!htb]
  \centering
  \includegraphics[width=0.85\textwidth]{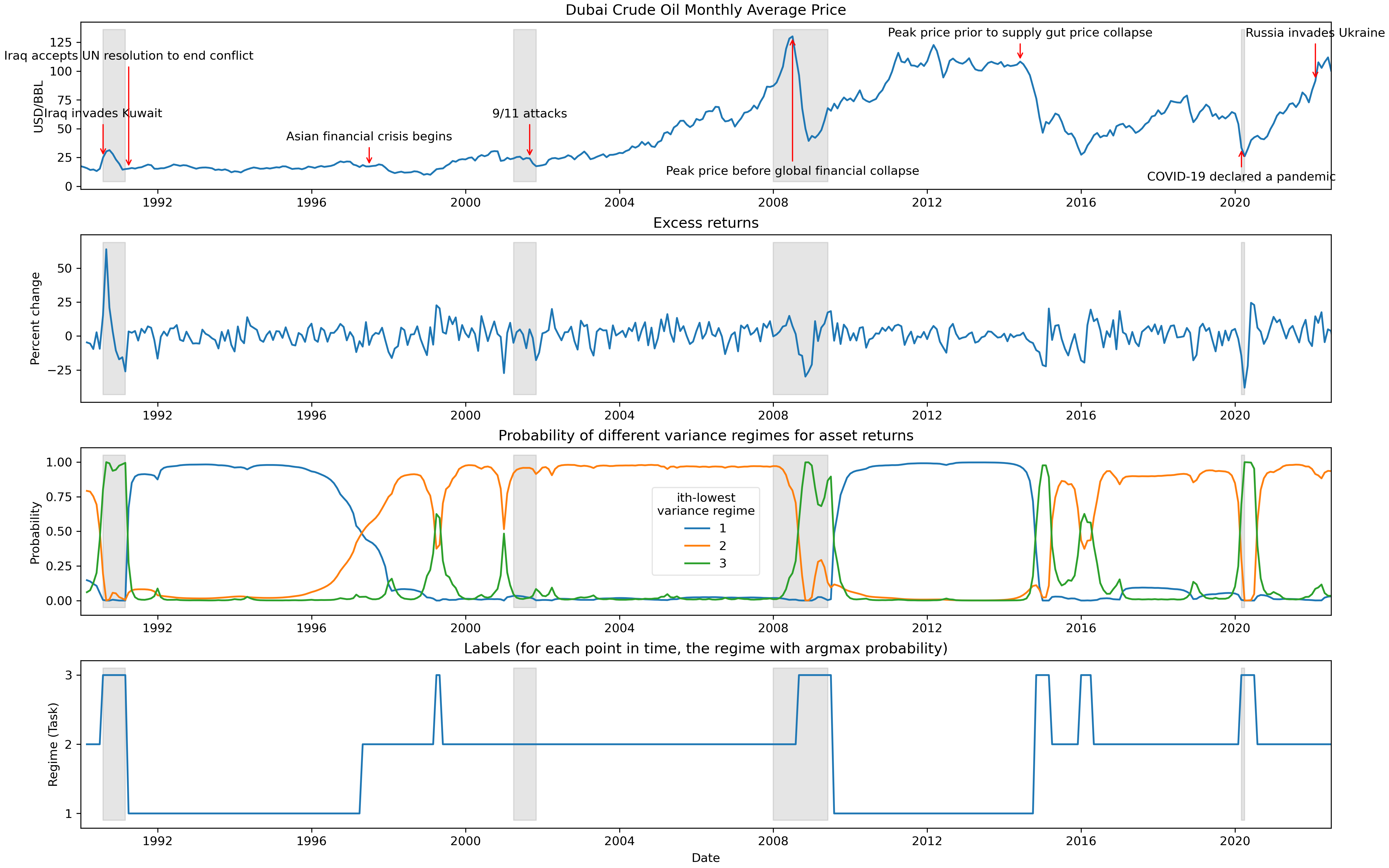}
  \captionof{figure}{\textit{(Top)} Dubai crude oil monthly average price per barrel. \textit{(Middle top)} Percent change over the \emph{monthly average price}, a proxy of volatility and variance, but not as granular as the weekly end price that the WTI and Brent datasets are using. \textit{(Middle bottom)} Fitted model regime probabilities over time. \textit{(Bottom)} Regime labels for each point in time, computed via argmax probabilites.}
  \label{fig:dubai_to_labels}
\end{figure*}

Also known as \textit{Fateh}, Dubai Crude is the least popular of the three primary global crude oil price benchmarks, but it is used as such because it is one of the few crude oils originating from the Persian Gulf that is available immediately. Dubai Crude has medium density and higher sulfur content than WTI and Brent but is priced comparably to both because of its central role in pricing crude oil exports from the Persian Gulf to Asian-Pacific markets. We depict our benchmark dataset creation algorithm for Dubai Crude in Figure \ref{fig:dubai_to_labels}

\section{Experimental Design and Results}
\label{sec:experiments}

We use the regime, or task, labels derived from our algorithm as context labels for four continual learning algorithms: MOLe \cite{nagabandi2018deep}, MoB \cite{pasula2023mixture}, MAML $k$-shot, and MAML-continuous. We evaluate on these algorithms because (1) each uses meta-learned priors $\theta^{*}$ for few-shot adaption to distribution shifts and (2) to cover a comprehensive set of continual learning algorithm /emph{classes}: a selector of expert models, a weighted mixture of expert and constructive non-expert models, a few-shot model that adapts from a history of $k$ data points, and a few-shot model that adapts continously from each new data point, respectively.

The first two, MOLe \cite{nagabandi2018deep} and MoB \cite{pasula2023mixture}, are modular algorithms, instantiating models from meta-learned priors when a new task is detected. In comparison to MOLe, a state-of-the-art approach for its time, MoB superceded MOLe by attaining both better overall performance and requiring fewer additional models to do so on multiple diverse domains while controlling for model architecture.

The latter two, MAML $k$-shot and MAML-continuous use a single model adapted from a meta-learned prior. MAML $k$-shot adapts using the $k$ latest data points, and MAML-continuous updates the model at each time step using the most recent observation. We chose $k=13$ for WTI and Brent and $k=3$ for Dubai Crude based on prior experience with similar time-series data and to align with the notion of \textit{quarterly} duration, which is of key importance in financial domains.

\begin{figure*}[ht]
  \centering
  \includegraphics[width=\textwidth]{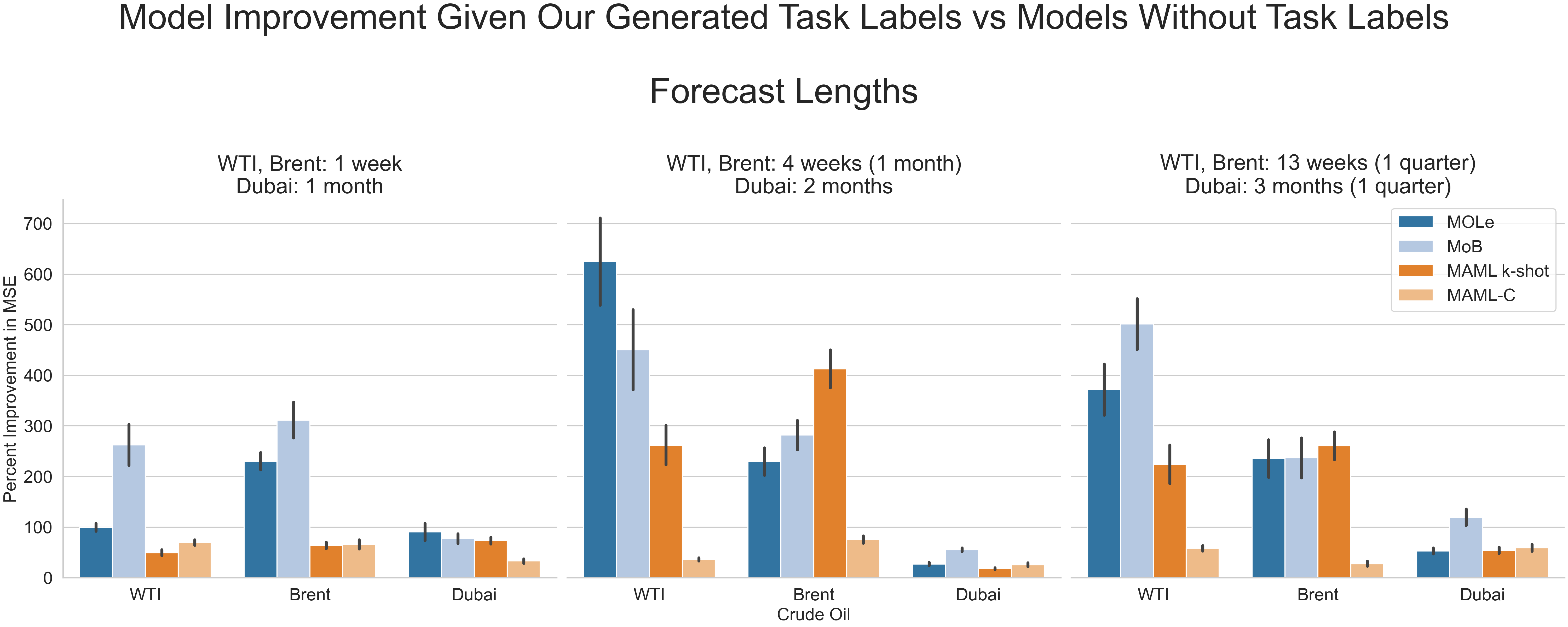}
  \captionof{figure}{The Percent Improvement in MSE after including our algorithm's generated contextual task labels as model inputs for each continual learning algorithm on each benchmark dataset over three financially and seasonally meaningful Forecasting Horizons. Each bar plot shows the mean and standard deviation "error" bars of 10 different seeds, so a total of 360 trials were conducted.}
  \label{fig:mse_imp}
\end{figure*}

We evaluate the usefulness of our contextual task labels by comparing the percent change in forecasting accuracy over financially and seasonally meaningful time-horizons with and without the inclusion of these task labels as model inputs. Since the original Dubai Crude time-series were average monthly prices, the horizons for this and for WTI and Brent differ. Results of the improvements after including the task labels are shown in Figure \ref{fig:mse_imp}.

By including our EM-based algorithm's automatically generated contextual task labels, we see universal improvement over each benchmark, time-horizon, and model considered. Since the raw data for Dubai Crude have already been averaged over month-long periods, we expect the performance changes on this benchmark to be relatively small compared to those of WTI and Brent, which is indeed what we see in Figure \ref{fig:mse_imp}.

\section{Limitations}
\label{sec:limitations}

While our work on time-series benchmarks in the financial domain provides valuable contributions, there are certain limitations to be acknowledged. These limitations include:

\emph{Domain Specificity}: The benchmark datasets we introduce focus solely on the financial domain, specifically crude oil prices. While this is an important domain, the applicability of our benchmarks to other domains may be limited.

\emph{Dataset Granularity}: The datasets we provide have varying levels of granularity, with daily spot prices for WTI and Brent and monthly average prices for Dubai Crude. This difference in frequency may impact the generalizability of our benchmarks across different time-series analysis tasks.

\emph{External Factors}: Our benchmarks consider major real-world events and contextual labels aligned with price and regime shifts. However, there may be additional external factors and events that could impact the performance of AI algorithms on the datasets, which are not explicitly captured in our work.

Addressing these limitations and expanding the scope of our benchmarks to encompass diverse domains and data granularities would be valuable directions for future research.
    
\section{Discussion}
\label{sec:discussion}
    In conclusion, we have introduced three novel time-series benchmark datasets that encompass the global crude oil market. These datasets were meticulously processed to fit into tasks using an expectation-maximization algorithm, providing valuable insight into temporal distribution shifts and their significant impact on predictive models. The benchmarks come equipped with labels aligning with significant real-world events, creating a rich context for further exploration and understanding. 
    
    Furthermore, we empirically show that including our algorithm's automatically generated task labels as model inputs universally improves performance over all benchmarks, time-horizons, and models considered.
    
    Our goal is to fuel advancements in areas such as continual learning and out-of-distribution detection, while addressing challenges posed by sequential data like non-stationarity and correlatedness. We believe that these benchmarks, due to the societal importance of the assets they represent, can make a substantial contribution to AI research, particularly in handling real-world data that contains distribution shifts. As AI continues to permeate every sector, the importance of such representative, real-world benchmarks will continue to grow.

\section*{Acknowledgments}
We would like to extend our sincerest gratitude to the IJCAI 2023 AI4TS Organizers and Program Committee. Their insightful reviews, constructive feedback, and recognition of the importance of this work have been instrumental in shaping the final manuscript. Their dedication to fostering innovation and excellence in AI is truly commendable, and we're honored that they have awarded this work in recognition of promoting these ideals.

\section*{Ethical Statement}

There are no ethical issues.

\bibliographystyle{named}
\bibliography{main}

\appendix
\label{app}

\section{Resampling Data}
\label{app:resamp}

\begin{figure}[!htb]
  \centering
  \includegraphics[width=\columnwidth]{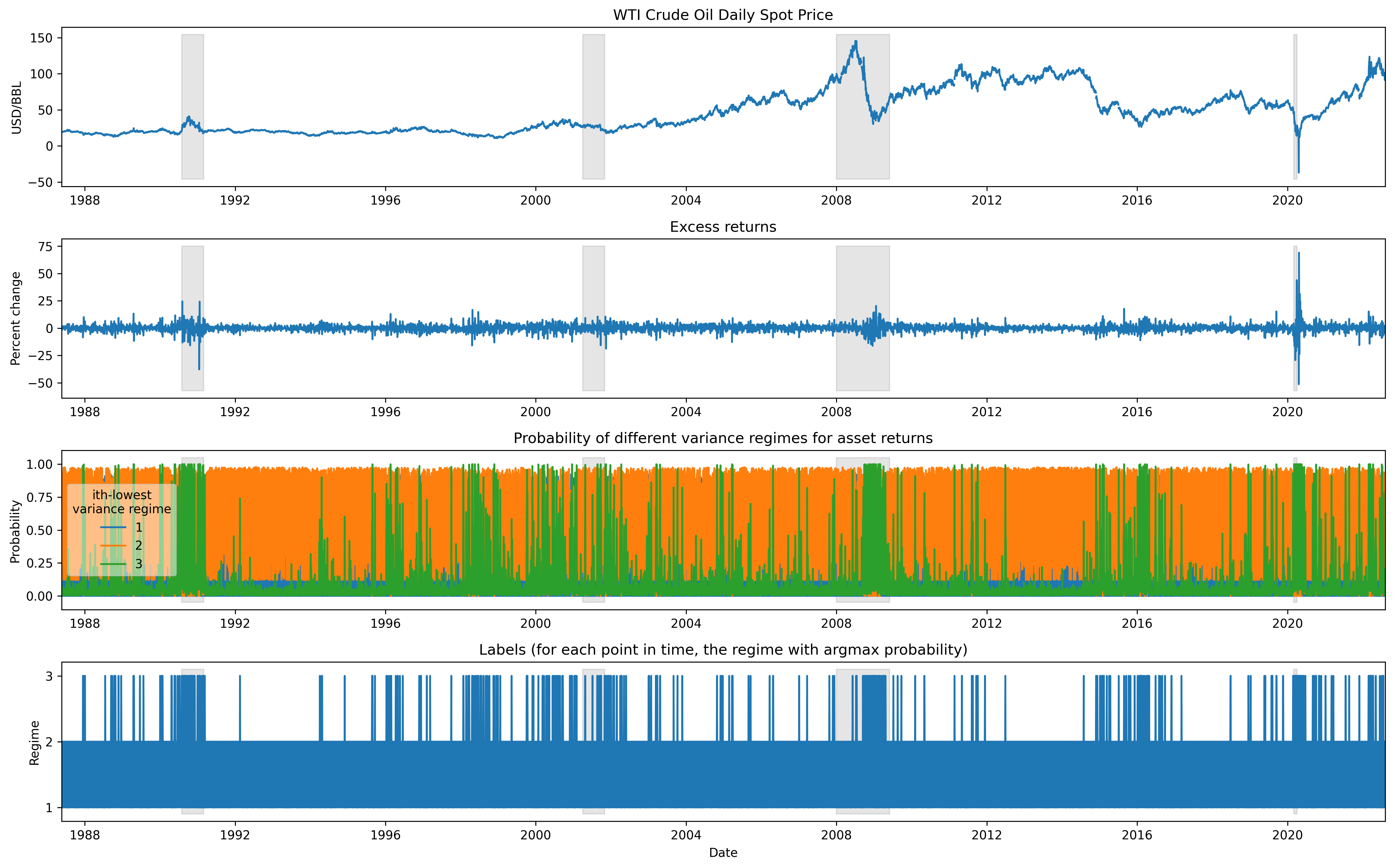}
  \captionof{figure}{Unrealistic task generation on WTI daily spot price data with a resampling frequency of every two days.}
  \label{fig:2dresamp}
\end{figure}

For daily frequency data, \citep{kim1998testing} returned unrealistic results, as mentioned in Section \ref{sssec:appropriate_resampling}. We resampled data by every two days, which produced similarly unrealistic results. For WTI, the results for this resampling frequency are shown in Figure \ref{fig:2dresamp}. We found that resampling weekly was an excellent frequency for \citep{kim1998testing} to generate tasks that were reasonable and well optimized measures of likelihood and information criteria. 

\section{Information Criteria}
\label{app:infocri}

We compute the AIC, BIC, HQIC, and the sum of these of the fitted model for $k = 2, 3, \dotsc, 10$ for each oil and show the results in Table \ref{tab:infocri}. Results for $k > 5$ are omitted because every model fit to this task count resulted in greater sum of information criteria than the minimum (in bold), had at least one task that didn't have argmax probability for any timesteps, and didn't converge. We ran the EM model fitting algorithm on each $k$ 200 times, and if no trial converged, then we list N under the row header Converge.

\begin{table}[!ht]
\centering
\scriptsize
\begin{tabular}{lcrrrrrc}\toprule
Oil &k Tasks &AIC &BIC &HQIC &Info Criteria Sum &Converge \\\midrule
WTI &2 &11025 &11047 &11033 &33104 &Y \\
&3 &10963 &11013 &10981 &\textbf{32957} &Y \\
&4 &10956 &11044 &10989 &32989 &N \\
&5 &10977 &11115 &11028 &33119 &N \\
\midrule
Brent &2 &10927 &10949 &10935 &32812 &Y \\
&3 &10850 &10899 &10868 &\textbf{32617} &Y \\
&4 &10860 &10949 &10893 &32702 &N \\
&5 &10882 &11020 &10933 &32834 &N \\
\midrule
Dubai &2 &2758 &2774 &2765 &\textbf{8297} &Y \\
&3 &2759 &2795 &2773 &8328 &Y \\
&4 &2773 &2837 &2799 &8409 &N \\
&5 &2790 &2889 &2830 &8509 &N \\
\bottomrule
\end{tabular}
\caption{Information criteria values for each oil for various $k$}
\label{tab:infocri}
\end{table}

Even though Dubai Crude's fitted model information criteria sum was minimized for $k=2$, since WTI and Brent fitted models are both minimized for $k=3$ and every task has argmax probability for at least one timestep for every oil for $k=3$, we use $k=3$ for every oil. 

Since Dubai Crude is sampled monthly, it's not surprising that a model with lesser $k$ than those for WTI and Brent, which are both resampled weekly, results in lesser information criteria sum.

\end{document}